\documentclass[10pt,twocolumn,letterpaper]{article}

\usepackage{cvpr}              
\usepackage{float}

\usepackage{algorithm}
\usepackage{algorithmic}
\usepackage{multirow, tabularx}
\usepackage{xtab} 
\usepackage{rotating}
\usepackage[dvipsnames]{xcolor}

\DeclareMathOperator*{\argminA}{arg\,min} 

\usepackage{amsthm}

\theoremstyle{plain}

\theoremstyle{definition}

\theoremstyle{remark}

\definecolor{cvprblue}{rgb}{0.21,0.49,0.74}
\usepackage[pagebackref,breaklinks,colorlinks,citecolor=cvprblue]{hyperref}

\def\paperID{*****} 
\def\confName{CVPR}
\def\confYear{2024}

\title{Discrete Cycle-Consistency Based Unsupervised Deep Graph Matching}

\author{Siddharth Tourani\textsuperscript{\rm 1,\rm 2}
Muhammad Haris Khan\textsuperscript{\rm 2},
Carsten Rother\textsuperscript{\rm 1},
Bogdan Savchynskyy\textsuperscript{\rm 1}\\
\textsuperscript{\rm 1}Computer Vision and Learning Lab, IWR, Heidelberg University\\
 \textsuperscript{\rm 2}Mohamed Bin Zayed University of Artificial Intelligence
}

\begin{document}
\maketitle
\begin{abstract}
We contribute to the sparsely populated area of unsupervised deep graph matching with application to keypoint matching in images. Contrary to the standard \emph{supervised} approach, our method does not require ground truth correspondences between keypoint pairs. Instead, it is \emph{self-supervised} by enforcing consistency of matchings between images of the same object category.
As the matching and the consistency loss are discrete, their derivatives cannot be straightforwardly used for learning. We address this issue in a principled way by building our method upon the recent results on black-box differentiation of combinatorial solvers. 
This makes our method exceptionally flexible, as it is compatible with arbitrary network architectures and combinatorial solvers. 
Our experimental evaluation suggests that our technique sets a new state-of-the-art for unsupervised deep graph matching.
\end{abstract}

\section{Introduction}
\label{sec:intro}

\begin{figure}[t]
  \centering
  \includegraphics[width=0.7\linewidth]{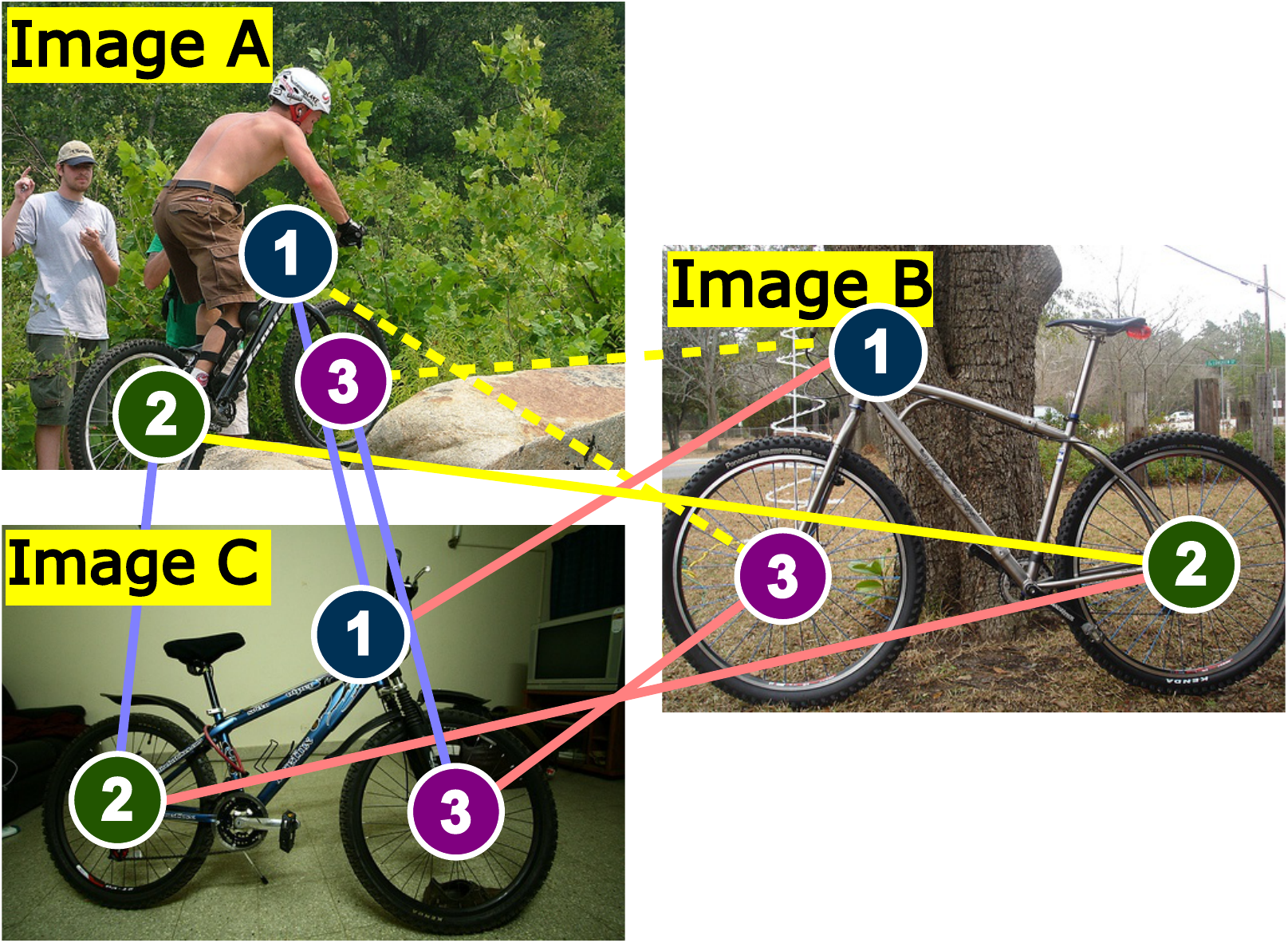}
   \caption{Illustration of cycle consistency in multi-graph matching (best viewed in color). There are three nodes in each image. They are labelled by both color (blue, green, purple) and numbers (1, 2, 3). Matches between pairs of nodes are shown by colored lines. $A \leftrightarrow B$, $B \leftrightarrow C$ and $C \leftrightarrow A$ are color coded with  yellow, light purple and light pink lines. Correct matches are shown by solid, wrong matches by dotted lines. Matching of the node 2 is cycle consistent across the images, whereas nodes 1 and 3 are not.}
   \label{fig:teaser}
\end{figure}

Graph matching (GM) is an important research topic in machine learning, computer vision, and related areas. It aims at finding an optimal node correspondence between graph-structured data. It can be applied in tasks like shape matching~\cite{sahilliouglu2020recent}, activity recognition~\cite{brendel2011learning}, point cloud registration~\cite{fu2021robust}, and many others. One classical application of graph matching also considered in our work is keypoint matching, as illustrated in \Cref{fig:teaser}.

A modern, learning-based approach to this problem tries to estimate costs for the subsequent combinatorial matching algorithm. The learning is usually supervised, \ie, ground truth correspondences are given as training data. However, obtaining ground truth is costly, which motivates development of unsupervised learning methods. 

Our work proposes one such unsupervised technique. Instead of ground truth correspondences, our method utilizes the \emph{cycle consistency} constraint as a supervision signal, see Fig.~\ref{fig:teaser}. Based on pairwise correspondences of multiple images, we iteratively update matching costs to improve consistency of the correspondences.

\subsection{Related Work}
\label{sec:related}
\paragraph{Supervised Deep Graph Matching} methods \cite{Fey/etal/2020,rolinek2020deep} typically consist of two parts: \emph{Feature extraction} and \emph{combinatorial optimization}. Whereas the first part is nowadays carried out by neural networks, the second is responsible for finding a one-to-one, possibly incomplete, matching. As shown in~\cite{poganvcic2019differentiation,battaglia2018relational}, neural networks do not generalize on combinatorial tasks and cannot substitute combinatorial methods therefore. 

The \textbf{architecture of neural networks} of recent deep graph matching methods such as~\cite{rolinek2020deep} or~\cite{wang2021neural} is very similar. As a backbone they use a VGG16~\cite{simonyan2014very} or a similar convolutional network for visual feature generation and a graph neural network for their refinement and combination with geometric information. Apart from specific parameters of the used networks, the key differences are the type of combinatorial solvers used and 
how differentiation through these solvers is dealt with.

A number of \textbf{combinatorial techniques} have been proposed to address the matching problem itself, see the recent benchmark~\cite{haller2022comparative} and references therein. One distinguishes between linear and quadratic formulations, closely related to the \emph{linear and quadratic assignment problems} (LAP and QAP resp.). These are deeply studied in operations research~\cite{burkard2012assignment}. Whereas the optimal linear assignment minimizes the sum of node-to-node (cf.\ keypoint-to-keypoint in Fig.~\ref{fig:teaser}) matching  costs, the quadratic assignment penalizes pairs of nodes (pairs of keypoints) matched to each other. Most importantly, this allows to take into account relative positions of respective keypoints. 

\begin{figure*}[t!]
\centering
\includegraphics[width=0.8\linewidth]{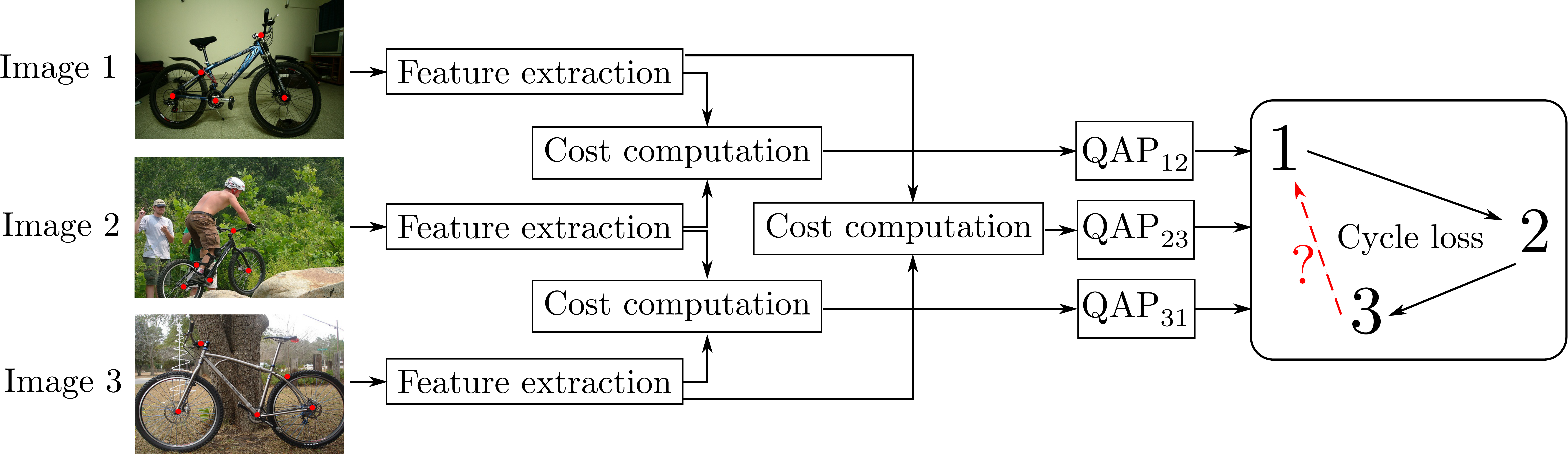}
\caption{Overview of our framework for a batch of 3 images. Features extracted from images and keypoint positions are transformed into matching costs for each pair of images. The QAP$_{ij}$ blocks compute the matching either as LAP or QAP. At the end the \emph{cycle loss} counts a number of inconsistent cycles and computes a gradient for back propagation.}
\label{fig:overview}
\end{figure*}

The greater expressive power of QAPs comes at a price: In general this problem is NP-hard, whereas LAP can be exactly and efficiently solved with a number of polynomial algorithms, see a practical analysis in~\cite{crouse2016implementing}. Most learning pipelines, however, adopt an approximate LAP solver based on the \emph{Sinkhorn normalization (SN)}~\cite{bregman1967relaxation,peyre2017computational} due to its inherent differentiability~\cite{eisenberger2022unified}. 
%

Despite the computational difficulty of QAPs, best existing algorithms are able to provide high-quality approximate solutions for problems with hundreds of keypoints within one second or even less~\cite{haller2022comparative}. This speed allows their direct use in end-to-end training pipelines provided there exists a way to differentiate through the solvers.

\textbf{The Differentiablity Issue.} 
When incorporating a combinatorial solver into a neural network, differentiability constitutes the principal difficulty. Such solvers take continuous inputs (matching costs in our case) and return a discrete output (an indicator vector of an optimal matching). This mapping is piecewise constant because a small change of costs typically does not affect the optimal matching. Therefore, the gradient exists almost everywhere but is equal to zero. This prohibits any gradient-based optimization.

The need for end-to-end differentiablity has given rise to various approaches that utilize differentiable relaxations of combinatorial optimization techniques~\cite{nowak2018revised, zanfir2018deep, jiang2019glmnet, wang2019learning, yu2019learning, fey2020deep, wang2021neural}. At the end of their pipeline, all these methods use the Sinkhorn normalization. \emph{However, such methods are either limited to approximately solving the LAP, or are solver specific, when addressing the QAP.} 

The most general technique providing a \emph{black-box differentiation of combinatorial solvers} was proposed in~\cite{poganvcic2019differentiation}. Applied to deep graph matching~\cite{rolinek2020deep} it is still among the state-of-the art methods in this domain. \emph{We also make use of this technique in our work, but in the unsupervised setting}.

\textbf{Multi-Graph Matching (MGM)} is a generalization of graph matching for computing correspondences between \emph{multiple} images of the same object. 
From an optimization point of view, the goal is to minimize the total matching cost between all pairs of such images given the \emph{cycle consistency} of all matchings~\cite{swoboda2019convex}. 
The problem is NP-hard even if the matching of each pair of objects is formulated as LAP~\cite{crama1992approximation,Tang2017coordMultiWay}. 
Apart from optimization, recent works in this domain include also learning of the matching costs, see, e.g.,~\cite{wang2021neural}. \emph{Contrary to these works, we do not enforce cycle consistency during inference, and only use it as a supervision signal for training.}

\textbf{Unsupervised Deep Graph Matching.}
The field of unsupervised deep graph matching is still under-studied. Essentially, it contains two works. The \emph{first} one~\cite{wang2020graduated}, referred to as GANN, uses cycle consistent output of an MGM solver as pseudo-ground truth for training a differentiable QAP-based matching.
Costs involved in MGM as well as the QAP are updated during training, to make the output of both algorithms closer to each other in the sense of a \emph{cross-entropy} loss.
The method is restricted to specific differentiable algorithms and biased to the sub-optimal solutions provided by the used MGM solver.

The \emph{second} unsupervised training technique called SCGM~\cite{liu2022self} is based on contrastive learning with data augmentation. Specifically, in the unsupervised training stage each image and the respective keypoint graph is matched to its augmented copies. The known mapping between original and modified keypoint graphs serves as ground truth. This technique can be applied with virtually any deep graph matching method as a backbone. Moreover, it can be used as a pre-training technique for our method, as demonstrated in \Cref{sec:experimental-validation}.


However, the augmentations are problem specific and dependent on an unknown data distribution. Also, SCGM uses two views of the same image to build its graph matching problem. It is thus biased towards complete matchings, which is a disadvantage in real world matching scenarios. 

\textbf{Cycle Consistency}
as self-supervision signal has been used in various computer vision applications such as videos, \eg,~\cite{wang2019learning} or dense semantic matching~\cite{kim2017fcss}. Another example is the seminal work~\cite{zhou2016learning} that leverages synthetic (3D CAD) objects to obtain correct (2D image-to-image) correspondences. In all these cases, however, one considers \emph{dense} image matching and penalizes the \emph{Euclidean or geodesic distance} between the first and the last point in a cycle. This type of loss does not fit the discrete setting, where nothing else but the number of incorrect matches has to be minimized, \ie, the Hamming distance between matches.

Another approach used, \eg, in multi-shape matching, \emph{implicitly} enforces cycle consistency by matching all objects to the \emph{universe}~\cite{ye2022joint}. This, however, eliminates cycle consistency as a supervision signal, hence one has to use additional information in the unsupervised setting. This is the functional map that delivers a supervision signal in the recent multi-shape matching works~\cite{cao2022unsupervised,cao2023unsupervised}. However, the lack of a functional map makes this approach inapplicable to graph matching in general and in our application in particular.

Finally, the recent work~\cite{indelman2023learning} uses a \emph{discrete} cycle consistency as part of a loss to improve the matching results in a \emph{supervised} setting. To differentiate the loss they use the \emph{direct loss minimization} technique~\cite{hazan2010direct} that can be seen as a limit case of the black-box-differentiation method~\cite{poganvcic2019differentiation} utilized in our framework.

\subsection{Contribution}
 We present a new principled framework for unsupervised end-to-end training of graph matching methods. It is based on a \emph{discrete} cycle loss as a supervision signal and the black-box differentiation technique for combinatorial solvers~\cite{poganvcic2019differentiation}. 
 Our framework can utilize \emph{arbitrary} network architectures as well as \emph{arbitrary} combinatorial solvers addressing LAP or QAP problems. It can handle incomplete matchings if the respective solver can. 

We demonstrate flexibility of our framework by testing it with two different network architectures as well as two different QAP and one LAP solvers. 

An extensive empirical evaluation suggests that our method sets a new state-of-the-art for unsupervised graph matching. 
Our code is available at: {\url{https://github.com/skt9/clum-semantic-correspondence}}.

\section{Overview of the Proposed Framework}
\Cref{fig:overview} provides an overview of our architecture. First the features are computed for each keypoint in each image (block \textbf{Feature extraction}). These features are translated to matching costs for each pair of images (block \textbf{Cost computation}) and the respective optimization problems, QAP or LAP (blocks \textbf{QAP$_{12}$}, \textbf{QAP$_{23}$}, \textbf{QAP$_{31}$}), are solved. Finally, \textbf{cycle loss} computes the \emph{number of inconsistent cycles}. 

We address all components one by one:\\
\begin{itemize}
\item \Cref{sec:background} overviews the black-box differentiation technique~\cite{poganvcic2019differentiation} that addresses the differentiability question discussed in~\Cref{sec:intro}. As the method requires a combinatorial solver to be represented in the integer linear program format, we briefly introduce this representation for LAP and QAP problems.
\item \Cref{sec:consistency-loss} describes the key component of our framework - the \emph{unsupervised discrete} cycle consistency loss and its differentiation.\\
\indent $\bullet$ In \Cref{sec:network-architecture} we propose a significant modification of the popular feature extraction and cost computation network of~\cite{rolinek2020deep}. The modified network is used in our experiments in addition to the original one.\\
\end{itemize}

%
%
\indent $\bullet$ Finally, our experimental validation is given in \Cref{sec:experimental-validation}. More detailed results are available in the supplement.

\begin{figure*}[h]
\centering
\includegraphics[width=0.9\textwidth]{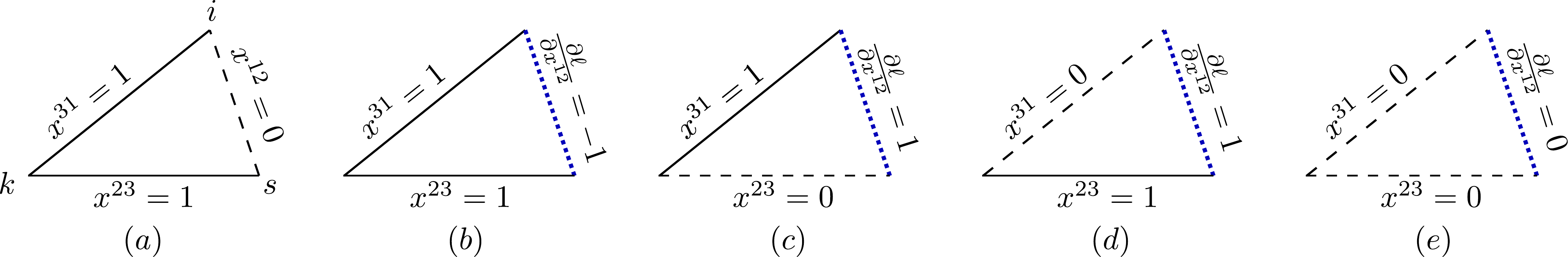}
\caption{(a) Partial loss illustration for a triple of indices $i,s,k$ and the respective binary variables $x^{12},x^{23},x^{31}$. The solid lines for $x^{23}$ and $x^{31}$ denote that these variables are equal to 1 and correspond to an actual matching between the respective points. The dashed line for $x^{12}$ denotes that this variable is equal to 0 and therefore points indexed by $i$ and $s$ are \emph{not} matched to each other. Given the values of $x^{23}$ and $x^{31}$ this violates cycle consistency. (b-e) Illustration of the values of the derivative $\partial{\ell}/\partial{x^{12}}$. 
The meaning of the solid and dashed lines as well as the position of $x^{12}$, $x^{23}$ and $x^{31}$ are the same as in (a).
The thick blue dotted lines mean that $x^{12}$ can be either 0 or 1, since $\partial{\ell}/\partial{x^{12}}$ is independent on $x^{12}$, see~\eqref{eqn:loss-gradient}.  So, for instance, $\partial{\ell}/\partial{x^{12}} = 1$ for $x^{23}=0$ and $x^{31}=1$ as illustrated by (c).}
\label{fig:loss-illustration}
\end{figure*}

\section{Background}
\label{sec:background}

\paragraph{Black-box differentiation of combinatorial solvers.}
The work~\cite{poganvcic2019differentiation} overcomes the zero gradient problem for discrete functions in a universal way. It introduces efficient piecewise-linear approximations of the considered piecewise-constant objective functions. It has the effect of making the gradient of the approximated function informative, thus allowing for gradient-based training. 
Let $\mathbf c \in \mathbb{R}^n$ be continuous input costs and $\mathbf x \in \mathcal{X}$ be a discrete output taken from an arbitrary finite set $\mathcal{X}$.
An important property of the method~\cite{poganvcic2019differentiation} is that it allows to use \emph{arbitrary} combinatorial solvers \emph{as a black-box}, as soon as the costs $\mathbf c$ and the output $\mathbf x$ are related via an \emph{integer linear program} (ILP):
\begin{equation}
\label{eqn:ilp-as-input-for-bbox}
    \mathbf x(\mathbf c) = \argminA_{\mathbf x \in \mathcal{X}}  \left\langle \mathbf c, \mathbf x\right\rangle\,.
\end{equation}
 This general formulation covers a significant portion of combinatorial problems including LAPs and QAPs. 

The flexibility of the black-box technique comes at the price of a somewhat higher computational cost. The used combinatorial solver must be run twice on each learning iteration: In addition to the normal execution on the forward pass, the backward pass involves another call to the solver with specially perturbed costs.

Essentially, if $L\colon\mathcal{X}\to\mathbb{R}$ denotes the final loss of the network, its gradient w.r.t. the costs $\mathbf c$ can be approximated as
\begin{equation}
\label{eqn:grad-interp}
    \frac{dL(\mathbf x(\mathbf c))}{d\mathbf{c}} := \frac{\mathbf x(\mathbf c^\lambda) - \mathbf x(\mathbf c)}{\lambda}\,.
\end{equation}
Here $\mathbf c^\lambda$ is a \emph{perturbed cost vector} computed as
\begin{equation}
    \mathbf c^\lambda = \mathbf c + \lambda \frac{dL}{d\mathbf x}(\mathbf x(\mathbf c))\,.
    \label{eqn:w-update}
\end{equation}
More precisely, \Cref{eqn:grad-interp} defines the gradient of a piecewise linear interpolation of $L$ at $\mathbf x(\mathbf c)$ with a hyperparameter $\lambda > 0$ controlling the interpolation accuracy.
%
\Cref{eqn:w-update} suggests that the loss function $L$ is differentiable. Note that for the gradient computation~\eqref{eqn:grad-interp} no explicit description of the set $\mathcal{X}$ is required.

\paragraph{Graph Matching Problem (QAP and LAP).}
For our description of the graph matching problem we mostly follow~\cite{haller2022comparative}.
Let $\mathcal{V}^1$ and  $\mathcal{V}^2$ be two finite sets to be matched, \eg, the sets of keypoints of two images. For each pair $i,j \in \mathcal{V}^1$, and each pair $s,l \in \mathcal{V}^2$, a cost $c_{is,jl}$ is given. Each such pair can be thought of as an edge between pairs of nodes of an underlying graph. In general, these costs are referred to as \emph{pairwise} or \emph{edge costs}. The \emph{unary}, or \emph{node-to-node} matching costs are defined by the diagonal terms of the \emph{cost matrix} $C=(c_{is,jl})$, \ie, $c_{is,is}$ is the cost for matching the node $i \in \mathcal{V}^1$ to the node $s \in \mathcal{V}^2$. For the sake of notation we further denote it as $c_{is}$.
In turn, $is$ will stand for $(i,s)$ below.

The goal of graph matching is to find a matching between elements of the sets $\mathcal{V}^1$ and $\mathcal{V}^2$ that minimizes the total cost over all pairs of assignments. It is represented as the following integer quadratic problem: 
\begin{align}
  \label{eqn:iqp}
  & \min_{\mathbf x \in \{0,1\}^{\mathcal{V}^1 \times \mathcal{V}^2}} \hspace{-3pt}
  \sum_{is \in \mathcal{V}^1 \times \mathcal{V}^2}\hspace{-10pt} c_{is} x_{is} 
  +\hspace{-10pt}\sum_{is,jl \in \mathcal{V}^1 \times \mathcal{V}^2 \atop is\neq jl}\hspace{-10pt} c_{is,jl} x_{is} x_{jl}\\
  & \text{s.t. } 
  \begin{cases}
  \forall i\in\mathcal{V}^1\colon\sum_{s\in\mathcal{V}^2}x_{is}\le 1\,,  \\
  \forall s\in\mathcal{V}^2\colon\sum_{i\in\mathcal{V}^1}x_{is}\le 1\,.
  \end{cases} \label{eqn:uniqueness-constraints}
\end{align}
The \emph{uniqueness constraints}~\eqref{eqn:uniqueness-constraints} specify that each node of the first graph can be assigned to \emph{at most one} node of the second graph. Due to the inequality in these constraints one speaks of \emph{incomplete} matching contrary to the equality case termed as \emph{complete}. The incomplete matching is much more natural for computer vision applications as it allows to account for noisy or occluded keypoints. From the computational point of view both problem variants are polynomially reducible to each other, see, \eg,~\cite{haller2022comparative} for details. Therefore we treat them equally unless specified otherwise.

By ignoring the second, quadratic term in~\eqref{eqn:iqp}, one obtains the LAP. Note that it already has the form~\eqref{eqn:ilp-as-input-for-bbox} required for black-box differentiation.
As for the more general QAP case, the substitution $y_{is,jl}=x_{is} x_{jl}$ linearizes the objective of~\eqref{eqn:iqp} and makes it amenable to black-box differentiation. The resulting ILP problem with different linearizations of the substitution $y_{is,jl}=x_{is} x_{jl}$ added as a constraint to the feasible set is addressed by a number of algorithms, see, \eg,~\cite{haller2022comparative}. We mention here only two, which we test within our framework: The LPMP solver~\cite{swoboda2017study} employed in the work~\cite{rolinek2020deep}, and the \emph{fusion moves} solver~\cite{hutschenreiter2021fusion} showing superior results in~\cite{haller2022comparative}.

\section{Cycle Consistency Loss and Its Derivative}\label{sec:consistency-loss}

\begin{algorithm*}[tb]
\caption{Unsupervised training algorithm}
  \label{alg:learning}
  \begin{algorithmic}
    \STATE \textbf{Given:} Sets of keypoints to be matched $\mathbb V:=\{ \mathcal V^i, i \in \{1,\dots,d\} \}$; $\lambda$ - the hyper-parameter from~\Cref{sec:background}.
    \STATE 1. Randomly select 3 sets from $\mathbb V$. W.l.o.g. assume these are $\mathcal V^1$, $\mathcal V^2$ and $\mathcal V^3$.
    \STATE 2. Infer costs $\mathbf c^{12},\mathbf c^{23},\mathbf c^{31}$ for the 3 QAPs corresponding to the pairs $(\mathcal V^1,\mathcal V^2)$, $(\mathcal V^2,\mathcal V^3)$, $(\mathcal V^3,\mathcal V^1)$.
    \STATE 3. Solve the QAPs and obtain the respective matchings $\mathbf x^{12}$, $\mathbf x^{23}$, $\mathbf x^{31}$, see~\Cref{fig:loss-illustration}.
    \STATE 4. Compute the perturbed costs $[\mathbf c^{12}]^\lambda$, $[\mathbf c^{23}]^\lambda$ and $[\mathbf c^{31}]^\lambda$ based on~\eqref{eqn:w-update} and the loss gradient~\eqref{eqn:loss-gradient}, \eg.:  
    \[
    [\mathbf c^{12}_{is}]^\lambda:= \mathbf c^{12}_{is} + \lambda \frac{\partial{L}}{\partial \mathbf x^{12}_{is}}\,,\ is\in\mathcal V^{1}\times\mathcal V^{2}\,.
    \]
        {
       \STATE 5. Compute the solutions $\mathbf x^{12}([\mathbf c^{12}]^\lambda)$, $\mathbf x^{23}([\mathbf c^{23}]^\lambda)$ and $\mathbf x^{31}([\mathbf c^{31}]^\lambda)$ to the QAP problems~\eqref{eqn:iqp} with the perturbed costs. } 
       \STATE 6. Compute the gradients via~\eqref{eqn:grad-interp} and backpropagate the changes to the network weights.
  \end{algorithmic}
\end{algorithm*}

Let us denote the fact that a point $i$ is matched to a point $s$ as $s \leftrightarrow i$. We call a mutual matching of $d$ point sets $\mathcal V^1,\dots,\mathcal V^d$ \emph{cycle consistent}, if for any matching sequence of the form $s^{k_1}\leftrightarrow s^{k_2}$, $s^{k_2}\leftrightarrow s^{k_3}$, $\dots$, $s^{k_{m-1}}\leftrightarrow s^{k_m}$ it holds $s^{k_m}\leftrightarrow s^{k_1}$, where $k_i\in \{1,\dots,d\}$, $k_i < k_{i+1}$, $i=1,\dots,m$.

It is well-known from the literature, see, \eg,~\cite{swoboda2019convex}, the cycle consistency over arbitrary subsets of matched point sets is equivalent to the cycle consistency of \emph{all triples}. We employ this in our pipeline, and define the cycle consistency loss for triples only.

Let us consider a matching of three sets $\mathcal V^1,\mathcal V^2$ and $\mathcal V^3$. 
We define the total cycle loss as the sum of \emph{partial} losses for all possible triples of points from these sets: $L(\mathbf x^{12},\mathbf x^{23},\mathbf x^{31})=\sum_{i\in\mathcal V^1}\sum_{s\in\mathcal V^2}\sum_{k\in\mathcal V^3}\ell(x^{12}_{is},x^{23}_{sk},x^{31}_{ki})$. 
Here $\mathbf x^{12}$ denotes a binary matching vector, \ie, a vector that satisfies the uniqueness constraints~\eqref{eqn:uniqueness-constraints} between $\mathcal V^{1}$ and $\mathcal V^{2}$. Vectors $\mathbf x^{23}$ and $\mathbf x^{31}$ are defined analogously.

Assume now that the triple of point indices $(i,s,k) \in\mathcal V^1\times\mathcal V^2\times\mathcal V^3$ is fixed. For the sake of notation we omit the lower indices and assume $x^{12}=x^{12}_{is}$, $x^{23}=x^{23}_{sk}$, $x^{31}=x^{31}_{ki}$ to be binary variables, see~\Cref{fig:loss-illustration}(a) for illustration.

The partial loss $\ell$ penalizes cycle inconsistent configurations as the one illustrated in~\Cref{fig:loss-illustration}(a). In particular, the partial loss is equal to $1$, if $x^{12}=0$ and $x^{23}=x^{31}=1$. This can be achieved by, \eg, the following differentiable function
\begin{multline}
\ell(x^{12},x^{23},x^{31})=\\(1-x^{12})x^{23}x^{31}+(1-x^{23})x^{12}x^{31}+(1-x^{31})x^{12}x^{23}\\
 = x^{12}x^{23}+x^{23}x^{31}+x^{12}x^{31}-3x^{12}x^{23}x^{31}
\end{multline}
where the three terms are necessary to make sure the loss function is symmetric. 

The derivative of the partial loss $\ell$ w.r.t.\ $x^{12}$ reads 
\begin{equation}\label{eqn:loss-gradient}
\frac{\partial{\ell}}{\partial x^{12}} = x^{23}+x^{31}-3x^{23}x^{31}\,,
\end{equation}
and analogously for variables $x^{23}$ and $x^{31}$.

\Cref{fig:loss-illustration}(b-e) illustrate the values of the derivative for the four possible cases. The gradient of $L$ is the sum of gradients of $\ell$ over all index triples. 

\Cref{alg:learning} summarizes our cycle-loss based unsupervised learning approach. Note that in Step 4 only unary costs $c_{is}$ are perturbed, as the pairwise costs $c_{is,jl}$ are multiplied by the lifted variables $y_{is,jl}$ in the linearized QAP objective, and $\partial L/\partial \mathbf y=0$.

\section{Network Architecture}\label{sec:network-architecture}
In order to show flexibility of our framework we tested it not only with the baseline network of~\cite{rolinek2020deep}, but also with our own network, whose architecture is presented in this section.
\subsection{Feature Extraction}  
\begin{figure*}[tb]
\setlength{\textfloatsep}{3pt}
\centering
    \includegraphics[scale=0.22]{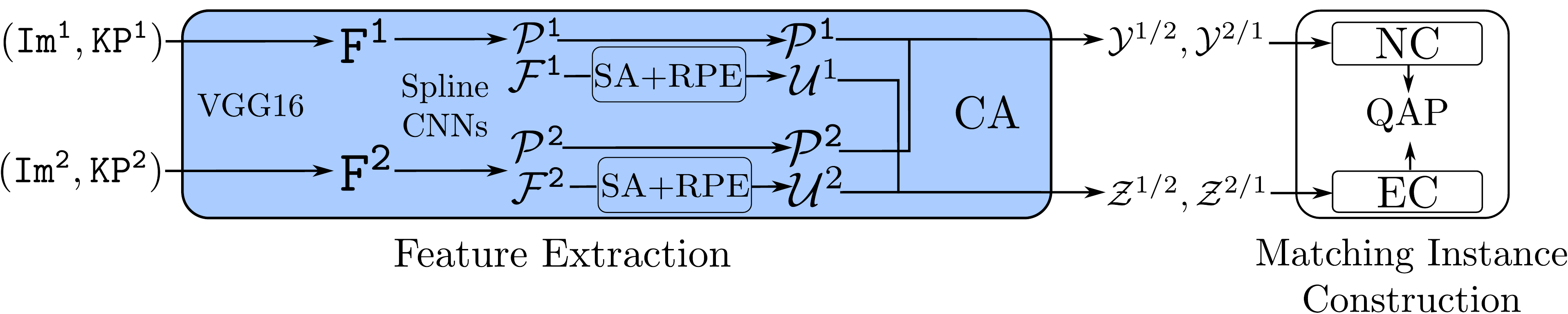}
    \caption{Information flow for feature processing and  matching instance construction. The feature extraction layer is shown in the blue box. Input to the pipeline are image-keypoint pairs, $(\texttt{Im}^1,\texttt{KP}^1)$, $(\texttt{Im}^2,\texttt{KP}^2)$ in the figure. The features extracted via a pre-trained VGG16 backbone network are refined by SplineCNN layers. The outputs of the SplineCNN layers are subsequently passed through self-attention (SA) with relative position encoding (RPE) and cross-attention (CA)layers and finally used in the construction of a matching instance. NC and EC denote node and edge costs. See the  detailed description in the main text.}
    \label{fig:feat-extract}
\end{figure*}

\Cref{fig:feat-extract} shows the information flow from input, through the feature extraction block, to the construction of the matching instance. The weights for the VGG16, SplineCNN and attention layers are shared across images. Input to the feature extraction block are the image-keypoint pairs. We denote them as $(\texttt{Im}^\texttt{1}, \texttt{KP}^\texttt{1})$ and $(\texttt{Im}^\texttt{2}, \texttt{KP}^\texttt{2})$.

The SplineCNN layers require a graph structure for the keypoints in each image. The keypoints form the nodes of the graph. We use the Delaunay Triangulation~\cite{delaunay6sphere} of the keypoint locations to define the edge structure of this graph. 
We refer to the graphs of $\texttt{Im}^1$ and $\texttt{Im}^2$ as $(\mathcal{V}^1, \mathcal{E}^1)$ and $(\mathcal{V}^2, \mathcal{E}^2)$, respectively. For the sake of notation, we denote an edge $(i,j)$ in an edge set $\mathcal{E}$ as $ij$.


\paragraph{Backbone Architecture.} Following the works of~\cite{rolinek2020deep,fey2020deep,wang2019learning} we compute the outputs of the $\mathtt{relu4\_2}$ and $\mathtt{relu5\_1}$ operations of the VGG16 network~\cite{simonyan2014very} pre-trained on ImageNet~\cite{deng2009imagenet}. The feature vector spatially corresponding to a particular keypoint is computed via bi-linear interpolation. The set of feature vectors thus obtained are denoted as $\texttt{F}^1, \texttt{F}^2$ in~\Cref{fig:feat-extract}.

\paragraph{SplineCNN Based Feature Refinement.} The keypoint features extracted from the VGG16 network are subsequently refined via SplineCNN layers~\cite{fey2018splinecnn}. SplineCNNs have been shown to successfully improve feature quality in point-cloud~\cite{li2020shape} and other graph structure processing applications~\cite{verma2021dual}.


The VGG16 keypoint features $(\texttt{F}^1, \texttt{F}^2$ in~\Cref{fig:feat-extract}) are input to the SplineCNN layers as node features. The input edge features to the SplineCNNs are defined as the difference of 2D coordinates of the associated nodes. We use two layers of SplineCNN with \texttt{MAX} aggregations. 

The outputs of the SplineCNN layers are appended to the original VGG node features to produce the refined node features, denoted as $\mathcal{F}^1, \mathcal{F}^2$. 
The edge features $\mathcal{P}^k$, $k=1,2$, are computed as $\mathcal P^k_{ii'}:=\mathcal F^{k}_i-\mathcal F^{k}_{i'}$, $ii'\in \mathcal E^k$.

\paragraph{Self- and Cross-Attention Layers} have been recently explored in the graph matching context~\cite{liu2023joint}. Essentially, they implement the function $\text{CA}(\mathbf{g}_i^1,\mathbf{g}^2,\mathbf p^Q,\mathbf p^K,\mathbf p^V):=$
{\scriptsize
 \begin{equation}
    \hspace*{-2pt}\sum_{s \in \mathcal V_2} \hspace{-2pt}\text{softmax}\hspace{-1pt}\left(\frac{(\mathbf{g}_i^1\mathbf{W}^Q+\mathbf p^Q_{is}).(\mathbf{g}_s^2\mathbf{W}^K+\mathbf p^K_{is})^\top}{\sqrt{D}}\right)\hspace{-1pt} \cdot\hspace{-1pt} (\mathbf{g}^1_i\mathbf{W}^V+\mathbf p^V_{is})\,
 \label{eqn:cross-attention-general}   
 \end{equation}
 }
$\mathbf{W}^Q$, $\mathbf{W}^K$ and $\mathbf{W}^V$ are learned projection matrices. $D$ is the dimension of the feature vector $\mathbf f^1_i$. \textbf{Q}, \textbf{K}, \textbf{V} stand for Query, Key and Value respectively. Intuitively speaking, the projection matrices learn which features to take more/less notice of. The vectors $\mathbf p^Q, \mathbf p^K, \mathbf p^V$ are described below.


\paragraph{Self-Attention + RPE layer} combines the self-attention mechanism~\cite{vaswani2017attention} with the \emph{relative position encoding (RPE)}. The latter has been shown to be useful for tasks where the relative ordering or distance of the elements matters \cite{wu2021rethinking}.

The layer transforms the node features $\mathcal{F}^k =\{ \mathbf{f}^k_i \mid i \in \mathcal{V}^k \}$, $k=1,2$ into the improved features 
$\mathcal{U}^{k} = \{ \mathbf{u}_i^{k} \mid i \in \mathcal{V}^k \}$, $k=1,2$. These are computed as $\mathbf{u}_i^{k}=\text{CA}(\mathbf{u}_i^k,\mathbf{f}^k,\mathbf p^Q,\mathbf p^K,\mathbf p^V)$, $k=1,2$.
The vector $\mathbf p^Q=\{\mathbf p^Q_{is}\mid is\in \mathcal V_1\times\mathcal V_2\}$ is computed as 
\begin{equation}
\mathbf p^Q_{is}=\mathtt{MLP}(\mathtt{sineEmbed}(x_i-x_s))\,,
\label{eqn:rpe-vec}
\end{equation}
where $x_i$ and $x_s$ are the 2D image coordinates of the respective key points. Here sineEmbed($\cdot$) stands for the \emph{sinusoidal embedding} consisting of 20 frequencies uniformly sampled in $[0,2\pi]$, as commonly used in transformers~\cite{vaswani2017attention}, and MLP is a multi-layer perceptron. Vectors $\mathbf p^K$ and~$\mathbf p^V$ are computed by~\eqref{eqn:rpe-vec} as well and differ only by learned weights of the respective MLPs.

\paragraph{Cross-Attention Layer} incorporates feature information across graphs. It has been used in a number of applications like semantic correspondence~\cite{yu2021cofinet} and point cloud registration~\cite{wang2019deep} to improve feature expressivity from two different data sources.

Recall that the node features refined by the Self-Attention layer are $\mathcal{U}^k =\{ \mathbf{u}^k_i \mid i \in \mathcal{V}^k \}$, $k=1,2$. 
Then the \emph{node cross attention features for $\mathcal{U}^1$ with respect to $\mathcal{U}^2$} denoted as $\mathcal{Z}^{1/2} = \{ \mathbf{z}_i^{1/2} \mid i \in \mathcal{V}^1 \}$ 
are defined as $\mathbf{z}^{1/2}_i = \text{CA}(\mathbf{u}_i^1,\mathbf{u}^2,\mathbf 0,\mathbf 0,\mathbf 0)$. The respective matrices $W^Q$, $W^K$ and $W^V$ defining this mapping are trained independently for self- and cross-attention layers.

The cross-attention node features for \emph{ $\mathcal{U}^2$ with respect to $\mathcal{U}^1$} are computed analogously and denoted by $\mathcal{Z}^{2/1} = \{ \mathbf{z}_s^{2/1} \mid s \in \mathcal{V}^2\}$. 

Similarly we compute the \emph{cross-attention edge features} $\mathcal{Y}^{1/2}=\{\mathbf{y}^{1/2}_{ij} \mid ij\in\mathcal E^1)\}$ and $\mathcal{Y}^{2/1}=\{\mathbf{y}^{2/1}_{ij} \mid ij\in\mathcal E^2)\}$ by plugging the coordinates of the edge features $\mathcal{P}^1$ and $\mathcal{P}^2$ into~\eqref{eqn:cross-attention-general} instead of the node features $\mathcal U^1$ and $\mathcal U^2$.

\subsection{Matching Instance Construction} 
It remains to specify how the costs for the graph matching problems in~\Cref{eqn:iqp} are computed. 
The unary costs $c_{is}$ are computed as:
\begin{equation}
\label{eqn:unary}
\small
c_{is}:= \left\langle \frac{\mathbf{z}^{1/2}_i}{||\mathbf{z}^{1/2}_i||},\frac{\mathbf{z}^{2/1}_s}{||\mathbf{z}^{2/1}_s||} \right\rangle  - \hat c\,. 
\end{equation}
The constant $\hat c$ regulates the number of unassigned points, \ie, its larger positive values decrease this number and smaller increase. We treat $\hat c$ as a hyper-parameter.
The edge costs  $c_{is,jl}$ are given by:
\begin{equation}\label{equ:quadratic-costs}
\small
c_{is,jl}:= \left\langle \frac{\mathbf{y}^{1/2}_{ij}}{||\mathbf{y}^{1/2}_{ij}||}, \frac{\mathbf{y}^{2/1}_{sl}}{||\mathbf{y}^{2/1}_{sl}||} \right\rangle\,.
\end{equation}

\section{Experimental Validation}\label{sec:experimental-validation}
\paragraph{Compared Methods} We evaluate our framework in four settings. As a \emph{baseline}, we build it on top of the supervised BBGM method~\cite{rolinek2020deep}. That is, we reuse its network and the respective QAP solver~\cite{swoboda2017study} within our unsupervised framework. We refer to the respective algorithm as CL-BBGM. 
Using BBGM as a baseline allows us to compare our method directly to the competing SCGM method~\cite{liu2022self} that uses BBGM as a backbone.

Our \emph{second} setting is a modification of CL-BBGM, referred to as CL-BBGM (SCGM), where we start with the weights learned in unsupervised fashion by SCGM~\cite{liu2022self} with BBGM as a backbone.
Our \emph{third} setup is the network described in~\Cref{sec:network-architecture} paired with the state-of-the-art QAP solver~\cite{hutschenreiter2021fusion}. We term it as CLUM, which stands for \emph{\textbf{C}ycle-\textbf{L}oss-based Unsupervised Graph \textbf{M}atching}. Our \emph{fourth} algorithm, referred to as CLUM-L, is a variant of CLUM with a LAP solver in place of the QAP one. The edge costs generated by the network are ignored in this case.

We compare our method to the so far only existing  unsupervised methods GANN~\cite{wang2020graduated} and SCGM~\cite{liu2022self}. As mentioned in \Cref{sec:intro}, SCGM is not stand-alone and requires a supervised graph matching algorithm as a backbone. Following the original SCGM paper~\cite{liu2022self}, we show results with backbones BBGM and NGMv2. We also provide published results of several supervised methods for reference, see \Cref{tab:compiled-results}.

\paragraph{Experimental Setup.} All experiments were run on an Nvidia-A100 GPU and a 32 core CPU. All reported results are averaged over 5 runs. The hyper-parameters are the same in all experiments. We used Adam~\cite{kingma2014adam} with an initial learning rate of $2 \times 10^{-3}$ which is halved at regular intervals. The VGG16 backbone learning rate is multiplied by 0.01. We process batches of 12 image triplets. 
The hyper-parameter $\lambda$ from~\eqref{eqn:w-update} is set to 80. Hyper-parameter $\hat c$ from~\eqref{eqn:unary} for Pascal VOC (unfiltered) is set to 0.21 for SCGM w/BBGM, 0.257 for both CLUM  and CLUM-L, 0.329 for both CL-BBGM and CL-BBGM (SCGM), respectively. Note, that $\hat c$ is important only for the case of incomplete assignments, \ie, the Pascal VOC (Unfiltered) dataset in our experiments, see below. In other experiments a sufficiently large value of $\hat c$ has been used to assure complete assignments.
We use image flips and rotations as augmentations. 

\begin{figure*}[t!]
\setlength{\tabcolsep}{1pt}
\renewcommand{\arraystretch}{3.1}
\begin{tabular}{c c c c}
{\texttt{\scriptsize \textbf{GANN}}}    &
\raisebox{-0.5\height}{\includegraphics[width=0.28\textwidth]{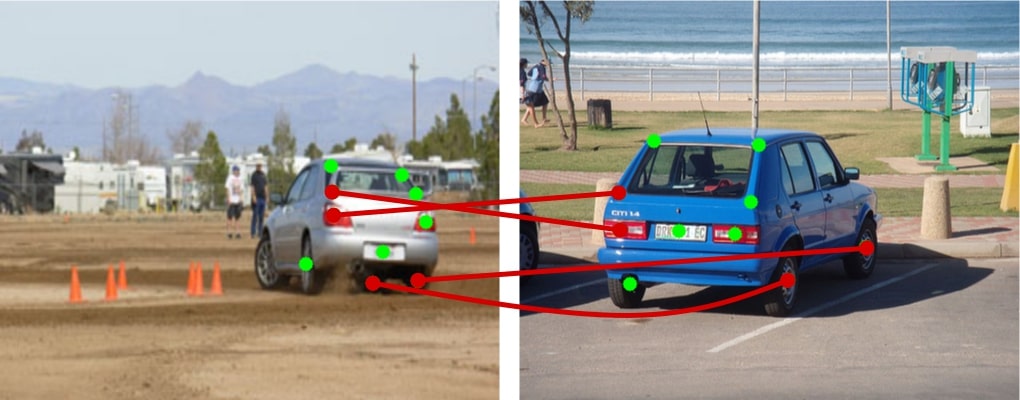}} & 
 \raisebox{-0.5\height}{\includegraphics[width=0.28\textwidth]{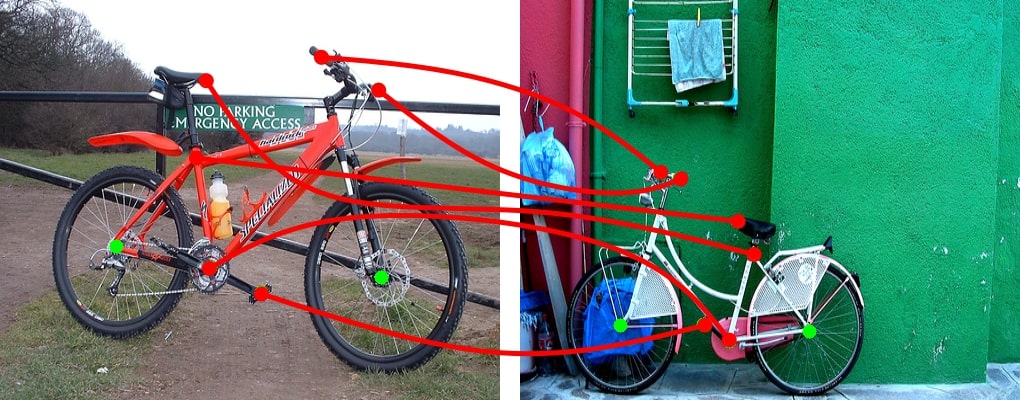}} & 
\raisebox{-0.5\height} {\includegraphics[width=0.28\textwidth]{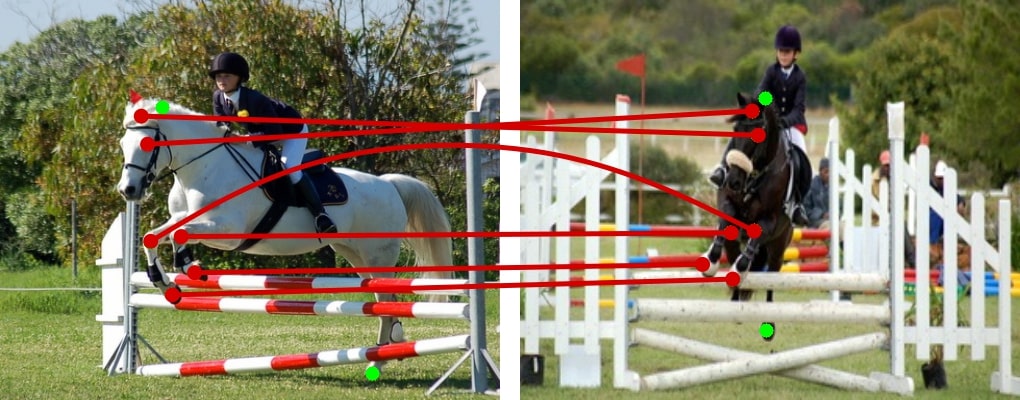}}\\

{\texttt{\scriptsize \textbf{CL-BBGM}}} &
\raisebox{-0.5\height}{\includegraphics[width=0.28\textwidth]{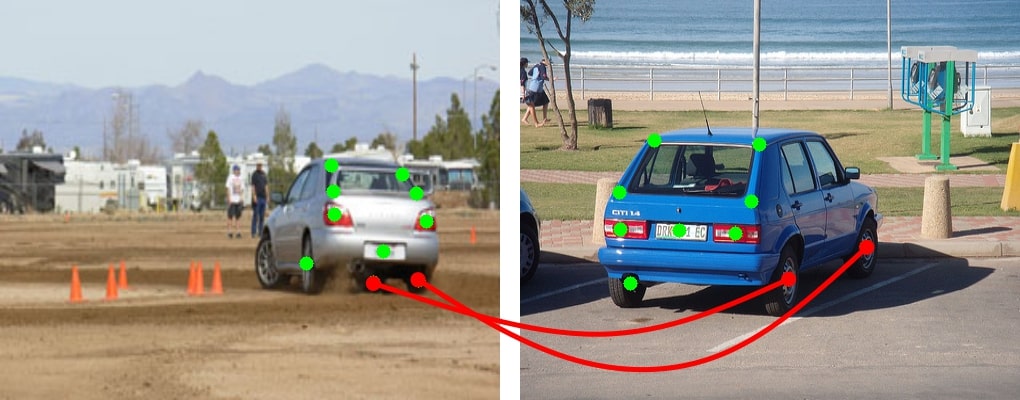}} &
\raisebox{-0.5\height}{\includegraphics[width=0.28\textwidth]{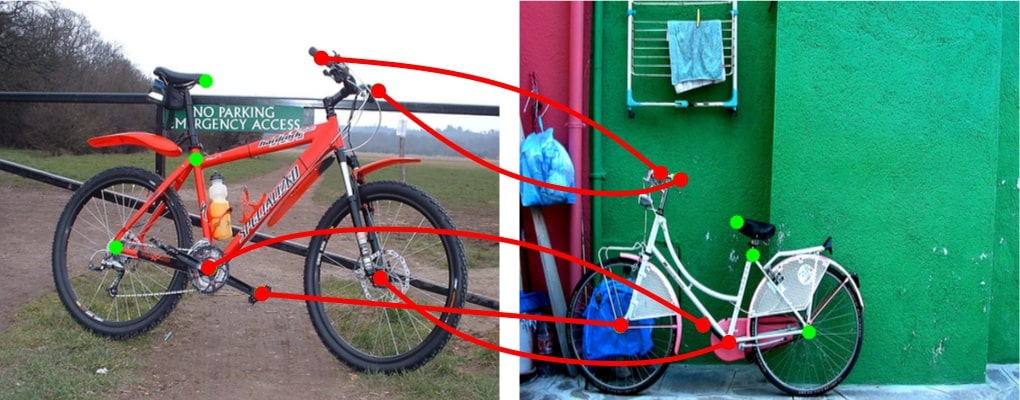}} &
\raisebox{-0.5\height}{{\includegraphics[width=0.28\textwidth]{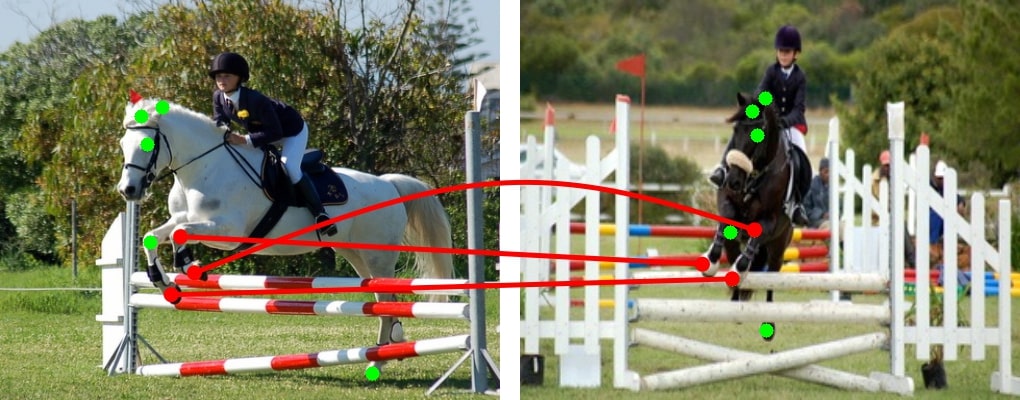}}}\\
{\texttt{\scriptsize \textbf{CLUM (Ours)}}} & 
\raisebox{-0.5\height}{\includegraphics[width=0.28\textwidth]{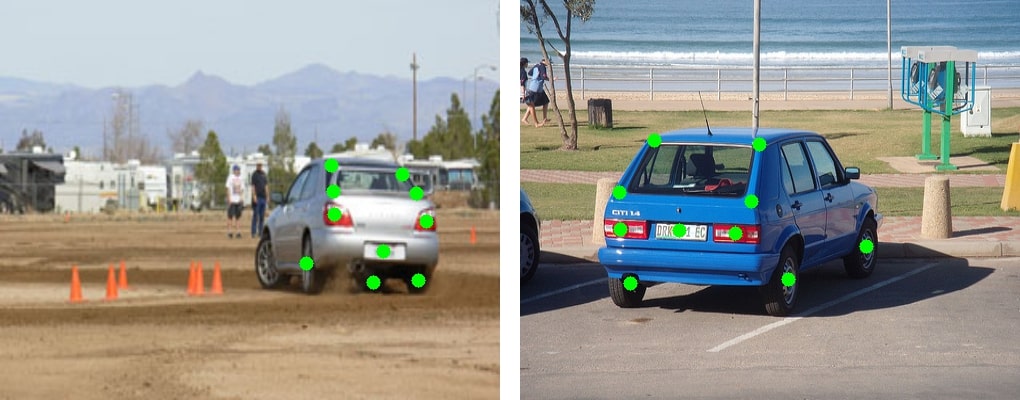}} & 
\raisebox{-0.5\height}{\includegraphics[width=0.28\textwidth]{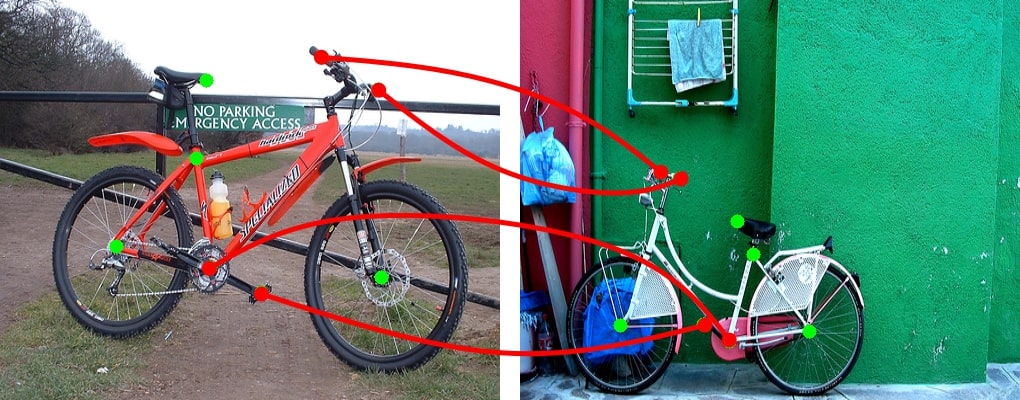}} &
\raisebox{-0.5\height}{\includegraphics[width=0.28\textwidth]{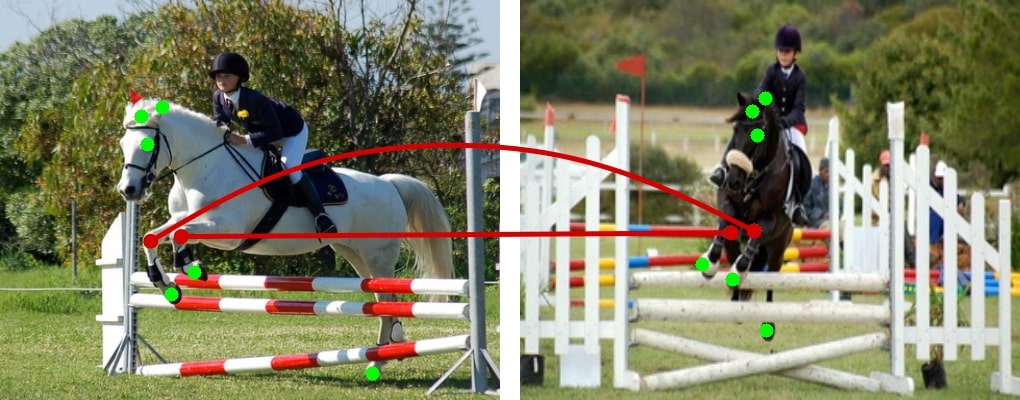}}\\
{\texttt{\scriptsize \textbf{BBGM (Supervised)}}} &
\raisebox{-0.5\height}{\includegraphics[width=0.28\textwidth]{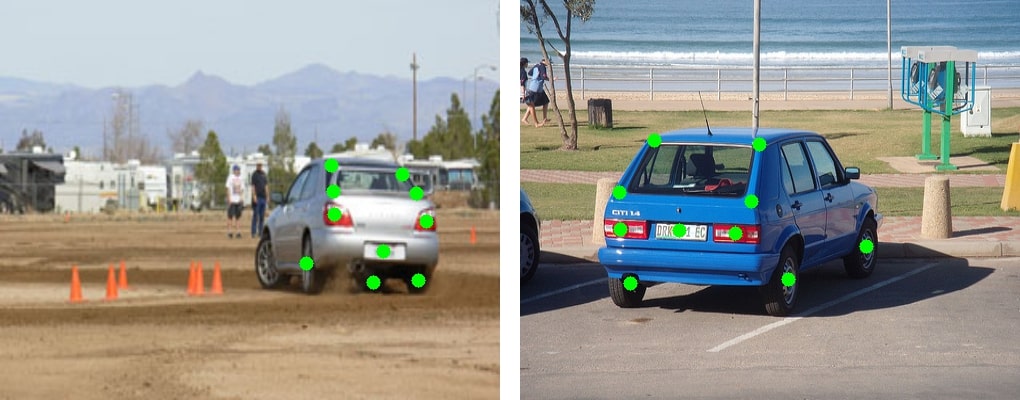}} &
\raisebox{-0.5\height}{\includegraphics[width=0.28\textwidth]{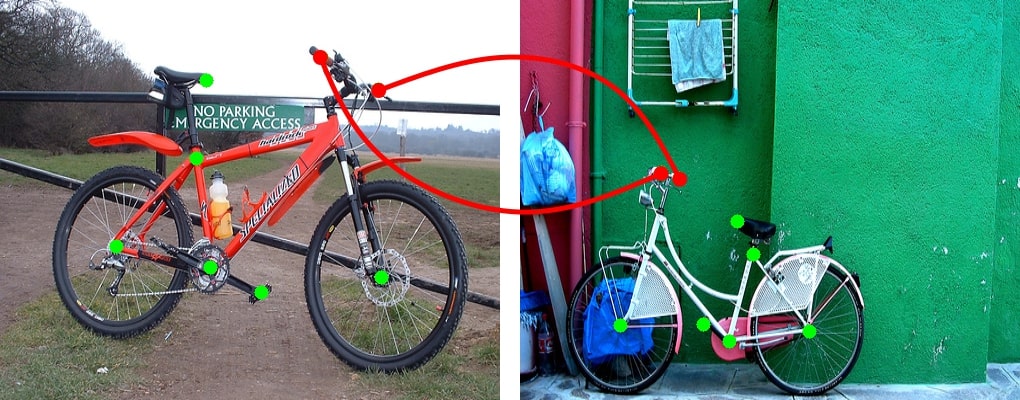}} &
\raisebox{-0.5\height}{\includegraphics[width=0.28\textwidth]{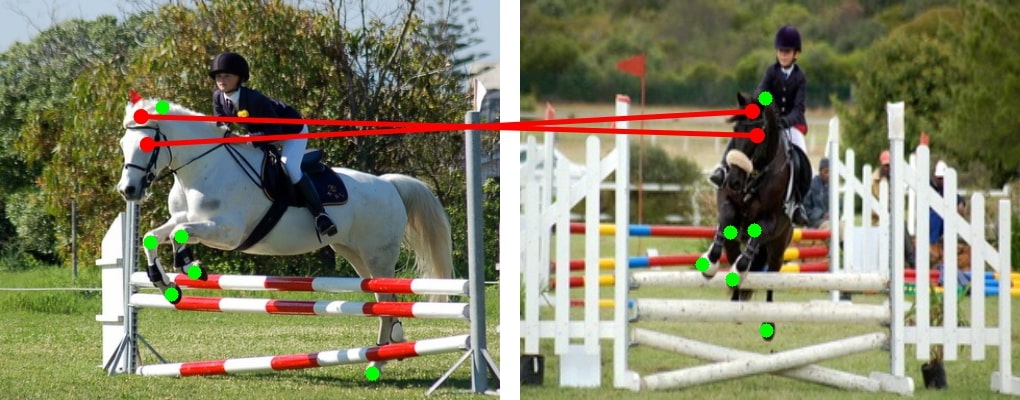}}\\
\end{tabular}
\caption{
Visualization of matching results on the SPair-71K dataset. In addition to the unsupervised techniques GANN, CL-BBGM, CLUM we show results of the fully supervised BBGM as a baseline.
Correctly matched keypoints are shown as green dots, whereas incorrect matches are represented by red lines.
The matched keypoints have in general similar appearance that suggests sensible unary costs. 
Improving of the matching quality from top to bottom is arguably mainly due to improving the pairwise costs, with the fully supervised BBGM method showing the best results.
}
\label{fig:spair-results}
\end{figure*}

\begin{table*}[t!]
\small
\begin{center}
\begin{tabular}{c|cc|cccccccc}
 & \multicolumn{2}{c|}{\textbf{Supervised}}& \multicolumn{7}{c}{\textbf{Unsupervised}} \\
   \textbf{Dataset} & BBGM & NGMv2  & GANN & SCGM & SCGM & CL-BBGM & CL-BBGM & \textbf{CLUM-L} & \textbf{CLUM}\\
   & & & & w/NGMv2 & w/BBGM  & & (SCGM) &  & \textbf{(Ours)}\\
   \midrule
PascalVOC (Filtered)  & \emph{79} & \emph{80.1} & \emph{31.5} & \emph{54.3} & \emph{57.1} &  58.4 & 58.8 &  59.7 &  \textbf{62.4}\\ 
    PascalVOC (Unfiltered) &  \emph{55.4} & \emph{54.0} &   \emph{24.3} & 32.1$^*$
 &  33.9$^*$ & 38 & 41.7  &  40.3 &  \textbf{43.5}\\ 
    Willow &  \emph{97.2} & \emph{97.5} &  \emph{92.0} &  \emph{91.0} & \emph{91.3}  & 
  91.6 & 93.2 &  93.4 &  \textbf{95.6}\\ 
    SPair-71K &  \emph{82.1} & \emph{80.2} & 31.7 & 36.9  & 38.7 & 40.6 & 
 41.2 & 41.6 & \textbf{43.1}\\
\end{tabular}
\caption{Consolidated results of the various deep graph matching methods on the benchmark datasets.
Numbers are accuracy in percentage (higher is better) for all datasets but PascalVOC (Unfiltered), where the F1-score is used.  
Detailed results can be found in the supplement. Italic font is used for the values taken from original works and the ThinkMatch library. \newline 
${\ }^*$ - trained on PascalVOC (Filtered).
}
\label{tab:compiled-results}
\end{center}
\end{table*}

\paragraph{Datasets} We evaluate our proposed method on the task of keypoint matching on the following datasets: Willow Object Class~\cite{cho2013learning}, SPair-71K~\cite{min2019spair} and Pascal VOC with Berkeley annotations~\cite{everingham2010pascal,bourdev2009poselets}. All but Pascal VOC assume \emph{complete} matching. The consolidated results are given in \Cref{tab:compiled-results}. The detailed evaluation can be found in~\cite{tourani2024unsupervisedGM-Arxiv}.

Following~\cite{rolinek2020deep} all considered methods are assumed to match pairs of images of the same category with at least three keypoints in common.
We apply the same rule to select the image triples for training. 

\paragraph{Pascal VOC with Berkeley Annotations.} 
%
 Following~\cite{rolinek2020deep}, we perform evaluations on the Pascal VOC  dataset in two regimes:\\
\indent $\bullet$\textit{Filtered.} 
Only the keypoints present in the matched images are preserved and all others are discarded as outliers. This corresponds to complete matching.\\
\indent $\bullet$ \textit{Unfiltered.} Original keypoints are used without any filtering. This corresponds to incomplete matching.

\paragraph{Willow Object Class.} 
Similar to other methods, we pre-train our method on Pascal VOC.

\paragraph{SPair-71K} is considered to have more difficult matching instances as well as higher annotation quality than PascalVOC.

\subsection{Results}

The \textbf{evaluation results} are summarized in~\Cref{tab:compiled-results} and illustrated in~\Cref{fig:spair-results}. 
In addition to the four unsupervised setups mentioned above, we trained GANN and SCGM on SPair-71K, as the respective results were missing in the original works. We also evaluated SCGM trained on PascalCOV (Filtered) on PascalVOC (Unfiltered), as  SCGM is not suitable for direct training for incomplete matching.
All other results are taken from the ThinkMatch~\cite{wang2021neural} testing webpage.


Note that already our baseline algorithm CL-BBGM outperforms all existing unsupervised methods (GANN and both SCGM variants) on all datastes but Willow, where it performs slightly worse than GANN. When pre-trained with SCGM (see CL-BBGM (SCGM) in \Cref{tab:compiled-results}) it gets consistently better results. In turn, our high-end setup CLUM uniformly outperforms all other unsupervised techniques. The LAP variant of this method, CLUM-L, performs significantly worse than CLUM, but still better than all previously existing unsupervised methods.

\subsection{Ablation Study}

\begin{table}[H]
\centering
\begin{tabular}{lcc}
Ablation & Pascal VOC & SPair-71K \\
\midrule
CLUM-RPE & 60.1 & 41.4\\
CLUM-RPE-Cross.~Att & 58.1 & 40.4\\
\midrule
CLUM  & 62.4 & 43.1 \\
CLUM-L & 59.7 & 41.6
\end{tabular}
\caption{The ablation study on the Pascal VOC (filtered) and SPair-71K datasets. Numbers show accuracy in percents.}
\label{tab:ablation}
\end{table}

 Results on PascalVOC (Filtered) and SPair-71K datasets are summarized in~\Cref{tab:ablation}. Here we sequentially remove the Self-Attention+RPE (CLUM-RPE) and Cross-Attention (CLUM-RPE--Cross.~Att) blocks that leads to monotonous degradation of the results quality.
In the CLUM-RPE experiment we set $\mathcal{S}^i:=\mathcal{F}^i$, for $i=1,2$.
In the CLUM-RPE--Cross.~Att experiment we substitute the cross-attention layer by setting 
$\mathbf{z}^{1/2}:=\mathcal F^1$, $\mathbf{z}^{2/1}:=\mathcal F^2$ and 
$\mathbf{y}^{1/2}:=\mathcal P^1$, $\mathbf{y}^{2/1}:=\mathcal P^2$.


\section{Conclusions}
We presented a new framework for unsupervised cycle-loss-based training of deep graph matching. It is extremely flexible in terms of the neural networks, as well as the combinatorial solvers it can be used with. Equipped with the best components it outperforms the state-of-the art and its flexibility suggests that its performance improves with the improvement of the components. Our framework can be adapted to other deep learning tasks like 6D pose estimation and correspondence estimation.

\section*{Acknowledgements}
We thank the Center for Information Services and High Performance Computing (ZIH) at TU Dresden for its facilities for high throughput calculations. Bogdan Savchynskyy was supported by the German Research Foundation (project number 498181230).

{
    \small
    \bibliographystyle{ieeenat_fullname}
    \bibliography{aaai24}
}

\newpage

\section*{Appendix: Detailed experimental result}
\label{sec:appendix}
\begin{minipage}{\textwidth}
\begin{table}[H]
\begin{center}
    \begin{tabular}{cccccccc}
    \multicolumn{8}{c}{\Large Willow Object Dataset}\\
        \\
\textbf{Type} & \textbf{Method} & Car & Duck & Face & Motorbike & Winebottle & \textbf{Mean}\\
\midrule
\multirow{4}{*}{\rotatebox[origin=c]{90}{ \small \textbf{Supervised}}} & \\
& BBGM & 96.8 & 89.9 & \textbf{100.0} & \textbf{99.8} & \textbf{99.4} & 97.2 \\
& NGM v2 & 97.4 & 93.4 & \textbf{100.0} & 98.6 & 98.3 & 97.5\\
& \\
\midrule
\multirow{7}{*}{\rotatebox[origin=c]{90}{\textbf{Unsupervised}}}  & \\
  & GANN & 85.4 & 89.8 & 100.0 & 88.6 & 96.4 & 92.0 \\
    & SCGM  w/ BBGM & 91.3 & 73.0 & 100.0 & 95.6 & 96.6 & 91.3 \\
    & SCGM  w/ NGM v2 &  91.2 & 74.4 & 99.7 & 96.8 & 92.7 & 91.0\\
    & CL-BBGM & 91.1 & 88.4 & 100.0 & 92.6 & 95.6 & 91.1\\
      & CL-BBGM (SCGM) & 92.2 & 76.8  & 100.0  & 97.1 & 95.9 & 91.4\\
   & CLUM-L & 93.1 & 94.2 & 100.0 & 97.3 &  97.6 & 94.1 \\
    & \textbf{CLUM (Ours)} & \textbf{94.7} &\textbf{95.7} & \textbf{100.0} & \textbf{97.8} & \textbf{98.6} & \textbf{95.6}\\
    & \\
    \midrule
    \end{tabular}    
\end{center}
\caption{Accuracy ($\%$) across all the object categories in the Willow Object dataset. Please note that all unsupervised methods are pretrained on Pascal VOC and fine-tuned on Willow Object.}
\label{tab:willow}
\end{table}
\end{minipage}

\begin{table*}[h!]
\begin{center}
    \begin{tabular}{ccccccccccccc}
    \multicolumn{12}{c}{\Large Pascal VOC (With Filtering)}\\
    \\
\textbf{Type} & \textbf{Method} & aero & bike & bird & boat & bottle & bus & car & cat & chair & cow  \\
\midrule
\multirow{4}{*}{\rotatebox[origin=c]{90}{\small \textbf{Supervised}}} & \\
& BB-GM & 61.9 & 71.1 & 79.7 & \textbf{79.0} & 87.4 & 94.0 & 89.5 & 80.2 & 56.8 & 79.1\\
& NGMv2 & 61.8 & 71.2 & 77.6 & 78.8 & 87.3 & 93.6 & 87.7 & 79.8 & 55.4 & 77.8 \\
 & \\
\midrule
\multirow{7}{*}{\rotatebox[origin=c]{90}{\small \textbf{Unsupervised}}} & \\
& GANN & 19.2 & 20.5 & 24.1 & 27.9 & 30.8 & 50.9 & 36.4 & 22.3 & 24.4 & 23.2 \\
 & SCGM w/BBGM & 37.6 & 49.9 & 54.8 & \textbf{54.5} & 65.6 & 56.4 & 60.6 & 52.3 & 36.8 & 51.4 \\
  & SCGM  w/ NGMv2 & 34.3 & 48.2 & 51.0 & 52.2 & 63.3 & 56.0 & 62.0 & 50.1 & \textbf{38.5} & 49.9 \\
    & CL-BBGM  & 38.3 & 51.4 & 57.7 & 53.1 & 67.4 & 54.9 & 62.3 & 51.4 & 35.1 & 52.1 \\
   & CL-BBGM (SCGM) & 38.9 & 52.5 & 58.1 & 53.4 & 67.9 & 55.3 & 63.3 & 52.3 & 36.9 & 52.9\\
   & CLUM-L  & 40.9 & 52.8 & 58.4 & 53.3 & 68.2 & 57.6 & 64.1 & 52.7 & 37.0 & 54.3 \\
    & \textbf{CLUM (Ours)} & \textbf{42.4} & \textbf{53.4} & \textbf{58.7} & 53.5 & \textbf{70.3} & \textbf{59.4} & \textbf{65.1} & \textbf{53.1} & 37.3 & \textbf{56.1}\\
    & \\
    \midrule
    & \\
 & \\
    \end{tabular}  
    \begin{tabular}{cccccccccccccc}
\textbf{Type} & \textbf{Method} & table & dog & horse & motor & person & plant & sheep & sofa & train & TV & \textbf{Mean} \\
\midrule
\multirow{4}{*}{\rotatebox[origin=c]{90}{\small \textbf{Supervised}}} & \\
& BB-GM &  64.6 & 78.9 & 76.2 & 75.1 & 65.2 & 98.2 & 77.3 & 77.0 & 94.9 & \textbf{93.9} & 79\\
& NGMv2 & 89.5 & 78.8 & 80.1 & 79.2 & 62.6 & 97.7 & 77.7 & 75.7 & 96.7& 93.2 & 80.1\\
 & \\
\midrule
\multirow{7}{*}{\rotatebox[origin=c]{90}{\textbf{Unsupervised}}} & \\
& GANN & 39.8 & 21.7 & 20.5 & 23.9 & 15.8 & 42.2 & 29.8 & 17.1 & 61.8 & 78.0 & 31.5\\
 & SCGM w/BBGM & 50.4 & 47.2 & 59.4 & 51.2 & 38.3 & \textbf{91.3} & 59.3 & 52.7 & 83.1 & \textbf{88.4} & 57.1\\
  & SCGM  w/ NGMv2 & 39.9 & 46.2 & 54.8 & 52.1 & 37.4 & 82.3 & 56.8 & 51.4 & 80.2 & 78.8 & 54.3\\
    & CL-BBGM  &  51.0 & 45.7 & 60.3 & 52.4 & 39.3 & 83.7 & 61.4 & 53.5 & 81.9 & 81.4 & 55.3 \\
    & CL-BBGM (SCGM) & 52.0 & 46.5 & 60.9 &  52.4 & 40.2 & 85.1 & 61.6 & 57.4 & 83.2 & 83.7 & 57.8\\
   & CLUM-L  & 53.5 & 47.95 & 62.4 & 53.1 & 42.0  & 86.2 & 61.9 &60.4  & 83.6 & 85.7 & 60.1 \\
    & \textbf{CLUM (Ours)}  & \textbf{55.0} & \textbf{49.4} & \textbf{63.6} & \textbf{53.8} & \textbf{43.8} & 87.3 & \textbf{62.1} &\textbf{63.4}& \textbf{83.9} & 87.6 & \textbf{62.4}\\
    & \\
    \midrule
    \end{tabular}    
\end{center}
\caption{Accuracy ($\%$) across all the object categories on Pascal VOC with intersection filtering.}
\label{tab:pascal-intersect}
\end{table*}

\begin{table*}[h!]
\begin{center}
    \setlength{\tabcolsep}{0.7em}
    \begin{tabular}{ccccccccccccc}
    \multicolumn{12}{c}{\Large Pascal VOC (Without Filtering)}\\
    &\\
\textbf{Type} & \textbf{Method}  & aero & bike & bird & boat & bottle & bus & car & cat & chair & cow  \\
\midrule
        &\\
\multirow{4}{*}{\rotatebox[origin=c]{90}{\textbf{Supervised}}} &\\
& BB-GM & 37.0 &  65.0 & \textbf{50.1} & \textbf{34.8} & 86.7 & \textbf{67.1} & 25.4 & \textbf{56.1} & \textbf{41.6} & \textbf{58.0} \\
& NGMv2 &  \textbf{39.4}  & \textbf{66.1} & 49.6 & 41.0 & \textbf{87.9} & 59.6 & \textbf{46.3} & 52.9 & 39.5 & 53.1 \\
& \\
& \\
\midrule
\multirow{7}{*}{\rotatebox[origin=c]{90}{\textbf{Unsupervised}}} & \\
& GANN & 12.6 & 19.5 & 16.6 & 18.5 & 41.1 & 32.4 & 19.3 & 12.3 & 24.3 & 17.2 \\
 & SCGM w/BBGM & 18.9 & 43.5 & 32.3 & 29.5 & \textbf{64.4} & 36.1 & 20.3 & 28.8 & 23.9 & 28.8 \\
  & SCGM  w/ NGMv2  & 19.7 & 42.2 & 29.5 & 23.9 & 62.3 & 35.2 & 21.2 & 27.3 & \textbf{24.1} & 25.9 \\
    & CL-BBGM & 21.4 & 43.8 & 29.5 & 23.9 & 61.7 & 37.8 & 21.9 & 30.2 & 23.5 & 29.1 \\
   & CL-BBGM (SCGM) & 21.6 & 46.0 & 30.9 & 27.1 & 62.6 & 38.8 & 23.2 & 30.3 & 26.5 & 32.4  \\
   & CLUM-L  & 22.6 & 44.5 & 26.3 & 26.7 & 63.4 & 36.1 &  19.8 & 31.9 & 21.4 & 33.5 \\
& \textbf{CLUM (Ours)} & \textbf{24.4} & \textbf{46.3} & \textbf{31.8} & \textbf{32.2} & 64.2 & \textbf{41.9} & \textbf{22.4} & \textbf{33.4} & 23.7 & \textbf{35.3} \\
& \\
    \midrule
    & \\
    & \\
    \end{tabular}

    \begin{tabular}{ccccccccccccc}
\textbf{Type} & \textbf{Method}  & table & dog & horse & motor & person & plant & sheep & sofa & train & TV & \textbf{Mean} \\
\hline
&\\
\multirow{4}{*}{\rotatebox[origin=c]{90}{\textbf{Supervised}}} &\\
& BB-GM  & \textbf{38.3} & \textbf{52.9} & \textbf{55.0} & \textbf{66.6} & 30.7 & \textbf{96.5} & \textbf{49.5} & 36.4 & \textbf{76.4} & \textbf{83.1} & \textbf{55.4}\\
& NGMv2 &   31.0 & 49.7 & 51.0 & 60.3 & \textbf{42.2} & 91.5 & 41.3 & \textbf{37.1} & 65.7 & 74.8 & 54.0\\
& \\
&\\
\hline
\multirow{7}{*}{\rotatebox[origin=c]{90}{\textbf{Unsupervised}}} & \\
& GANN &  38.0 & 12.2 & 15.9 & 18.2 & 19.4 & 35.5 & 14.8 & 15.4 & 41.5 & 60.8 & 24.3\\
 & SCGM w/BBGM &  23.7 & 23.3 & 31.4 & 33.4 & 21.1 & 83.2 & 25.5 & 27.0 & 49.4 & \textbf{72.9} & 35.9\\
  & SCGM  w/ NGMv2  & 22.8 & 23.5 & 30.3 & \textbf{35.7} & \textbf{21.3} & 67.5 & 24.6 & 21.6 & 44.4 & 65.6 & 33.3\\
    & CL-BBGM & 28.7 & 22.7 & 29.7 & 32.7 & 19.3 & 77.8 & 25.8 & 28.6 & 50.4 & 67.6 & 36.3\\
& CL-BBGM (SCGM) & 32.4 &  23.7 & 31.3 & 32.7 & 22.6 & 78.3  & 29.6 & 28.7 & 53.1 & 70.9  & 37.2\\
   & CLUM-L  & 40.9 & 24.2 & 
  33.2 & 32.7 & 22.0 &  82.9 & 25.8 & 28.5&
  53.7 & 67.1  & 40.9 \\
& \textbf{CLUM (Ours)} & \textbf{42.1} & \textbf{26.3} & \textbf{35.1} & 33.8 & 22.1 & \textbf{86.2} & \textbf{30.1} & \textbf{33.2} & \textbf{54.3} & 72.1 & \textbf{43.5}\\
& \\
    \hline
    \end{tabular}    
\end{center}
\caption{F1 score ($\%$) (the higher the better) across all the object categories on Pascal VOC without filtering.}
\label{tab:pascal-union}
\end{table*}

\begin{table*}[h!]
\begin{center}
    \begin{tabular}{ccccccccccccc}
    \multicolumn{12}{c}{\Large SPair-71K}\\
        \\
\textbf{Type} & \textbf{Method} & aero & bicycle & chicken & boat & bottle & bus & car & cat & chair & cow \\
\hline
\multirow{4}{*}{\rotatebox[origin=c]{90}{\small \textbf{Supervised}}} & \\
& BB-GM & \textbf{72.5} &  \textbf{64.6} & 87.8 & 75.8 & 69.3 & 93.9 & 88.6 & \textbf{79.9} & 74.6 & 83.2 \\
& NGMv2  &  68.8 & 63.3 & 86.8 & 70.1 & \textbf{69.7} & \textbf{94.7} & 87.4 & 77.4 & 72.1 & 80.7\\

& \\
\hline
\multirow{7}{*}{\rotatebox[origin=c]{90}{\small \textbf{Unsupervised}}}
& \\
& GANN  & 27.8 & 22.4 & 41.8 & 19.5 & 37.1 & \textbf{49.8} & 24.5 & 15.9 & 24.2 & 38.7 \\
    & CL-BBGM &  29.3 & 24.3 & 44.5 & 22.5 & 38.4 & 46.7 & 30.1 & \textbf{28.3} & 28.9 & 39.0 \\
   & CL-BBGM (SCGM) & 30.3 & 25.8 & 46.6 & 24.3 & 40.3 & 49.0 &
  32.0 & 29.3 & 30.9 & 40.1 \\
   & CLUM-L  & 30.8 &  27.2 & 43.4 &  25.5 & 37.6 & 47.2 & 28.6 & 26.8 & 32.4 & 42.2\\
    & \textbf{CLUM (Ours)} & \textbf{32.4}  & \textbf{27.7} & \textbf{47.2} & \textbf{26.3} & \textbf{40.6} & 48.7 & \textbf{30.9} & 27.7 & \textbf{34.3} & \textbf{43.9} \\    
    &   &  \\
    \hline
    &   &  \\
    \end{tabular}  

        \begin{tabular}{cccccccccccc}
        \\
\textbf{Type} & \textbf{Method}  & dot & horse & motor & person & plant & sheep & train & TV & \textbf{Mean}\\
\hline

\multirow{4}{*}{\rotatebox[origin=c]{90}{\small \textbf{Supervised}}} & \\
& BB-GM & \textbf{78.8} & \textbf{77.1} & 76.5 & 76.3 & 98.2 & \textbf{85.5} & \textbf{96.8} & 99.3 & 82.1 \\
& NGMv2  &   74.3 & 72.5 & \textbf{79.5} & 73.4 & 98.9 & 81.2 & 94.3 & 98.7 & 80.2\\
& \\
\hline
\multirow{7}{*}{\rotatebox[origin=c]{90}{\small \textbf{Unsupervised}}}
& \\
& GANN  & 23.9 & 17.3 & \textbf{29.3} & 17.6 & 40.3 & 19.9 & 56.6 & \textbf{64.9} & 31.7\\
    & CL-BBGM &  34.3 & 55.6 & 25.1 & 45.4 & 53.1 & 29.6 & 73.4 & 59.1 & 39.3\\
   & CL-BBGM (SCGM) & 34.4 & 58.6 & 26.9 & 47.7 & 54.7 & 29.7 & 73.6 & 59.5 & 40.6 \\
   & CLUM-L  & 40.53 & 59.0 & 25.9 & 44.7 & 50.5 & 30.4 & 77.4 & 61.9 & 41.0\\
    & \textbf{CLUM (Ours)} &  \textbf{40.7} & \textbf{62.1} & 27.8 & \textbf{46.1}& 54.1 & \textbf{34.1} & \textbf{79.4} & 64.2 & \textbf{43.1}\\    
    &   &  \\
    \hline
    \end{tabular}
\end{center}
\caption{Accuracy ($\%$) across all the object categories.}
\label{tab:spair}
\end{table*}

\end{document}